\documentclass[english]{article}
\usepackage[T1]{fontenc}
\usepackage[utf8]{inputenc}
\usepackage{geometry}
\usepackage{color}
\usepackage{url}
\usepackage{babel}
\usepackage{booktabs}
\usepackage{multirow}
\usepackage{siunitx}
\usepackage{mathtools}
\usepackage{bm}
\usepackage{bbm}
\usepackage{amsmath}
\usepackage{amsthm}
\usepackage{amssymb}
\usepackage[linesnumbered,ruled,vlined]{algorithm2e}
\usepackage{algorithmic}
\usepackage{graphicx}
\usepackage{caption}
\usepackage{subcaption}
\usepackage[unicode=true,pdfusetitle,
 bookmarks=true,bookmarksnumbered=true,bookmarksopen=true,bookmarksopenlevel=1,
 breaklinks=false,pdfborder={0 0 0},pdfborderstyle={},backref=false,colorlinks=true]
 {hyperref}
\hypersetup{
 linkcolor=blue, urlcolor=blue, citecolor=blue, linktoc=all, linktocpage=false,pdfauthor={Name}}
\hypersetup{bookmarksopen=true,
        bookmarksopenlevel=1, 
        bookmarksnumbered=true, }
\usepackage{appendix}

\usepackage[most]{tcolorbox}

\makeatletter

\theoremstyle{plain}

\theoremstyle{remark}
\newtheorem{remark}{\protect\remarkname}
\theoremstyle{plain}

\theoremstyle{plain}

\theoremstyle{plain}

\theoremstyle{plain}

\theoremstyle{plain}

\theoremstyle{definition}

\theoremstyle{plain}

\definecolor{todo}{RGB}{0,200,200}

\definecolor{new}{RGB}{0,200,200}
\newcommand{\new}[1]{\textcolor{todo}{[NEW:]}}

\definecolor{emerald}{rgb}{0.31, 0.78, 0.47}

\allowdisplaybreaks

\renewcommand{\Pr}{{\mathbb{P}}}

\newcommand{\R}{{\mathbb{R}}}

\newcommand{\bc}{\bm{c}}

\newcommand{\bx}{\bm{x}}

\newcommand{\btheta}{\bm{\theta}}

\newcommand{\Lcal}{{\mathcal{L}}}

\newcommand{\loss}{\Lcal}

\newcommand{\eetuning}{\texttt{EE-Tuning}\xspace} 
\newcommand{\eellm}{{EE-LLM}\xspace}

\newcommand{\llama}{Llama~2-Chat\xspace}
\newcommand{\archembed}{\texttt{Embedding}\xspace}
\newcommand{\archnorm}{\texttt{Norm}\xspace}
\newcommand{\archmlp}{\texttt{MLP}\xspace}
\newcommand{\archlayer}{\texttt{Layer}\xspace}
\newcommand{\pt}{pre-training\xspace} 
\newcommand{\ift}{IFT\xspace} 
\newcommand{\initcopy}{\texttt{copy}\xspace}
\newcommand{\initrandom}{\texttt{random}\xspace}

\makeatother

\providecommand{\assumptionname}{Assumption}
\providecommand{\corollaryname}{Corollary}
\providecommand{\examplename}{Example}
\providecommand{\factname}{Fact}
\providecommand{\lemmaname}{Lemma}
\providecommand{\propositionname}{Proposition}
\providecommand{\remarkname}{Remark}
\providecommand{\theoremname}{Theorem}

\begin{document}
\title{\eetuning: An Economical yet Scalable Solution for Tuning\\Early-Exit Large Language Models}

\author{
Xuchen Pan$^*$,
Yanxi Chen$^*$,
Yaliang Li,
Bolin Ding,
Jingren Zhou \\
\small{\texttt{\{panxuchen.pxc, chenyanxi.cyx, yaliang.li, bolin.ding, jingren.zhou\}@alibaba-inc.com}}\\
Alibaba Group
}

\date{}

\maketitle

\renewcommand*{\thefootnote}{\fnsymbol{footnote}}
\footnotetext[1]{Co-first authors. } 
\renewcommand*{\thefootnote}{\arabic{footnote}}

\begin{abstract}

This work introduces \eetuning, a lightweight and economical solution to training/tuning \emph{early-exit} large language models (LLMs).
In contrast to the common approach of full-parameter pre-training, 
\eetuning augments any pre-trained (and possibly fine-tuned) standard LLM with additional early-exit layers that are tuned in a parameter-efficient manner, 
which requires significantly less computational resources and training data.
Our implementation of \eetuning achieves outstanding training efficiency via extensive performance optimizations,
as well as scalability due to its full compatibility with 3D parallelism.
Results of systematic experiments validate the efficacy of \eetuning,
confirming that effective early-exit LLM inference can be achieved with a limited training budget.
In hope of making early-exit LLMs accessible to the community, 
we release the source code of our implementation of \eetuning at \url{https://github.com/pan-x-c/EE-LLM}.

\end{abstract}

\setcounter{tocdepth}{2}
\tableofcontents

\section{Introduction}

Transformer-based large language models (LLMs) have achieved extraordinary performance on various language tasks \cite{vaswani2017attention,Brown2020,OpenAI2023,Touvron2023llama,Touvron2023llama2,Chowdhery2022}.
Meanwhile, these models incur high costs and latency during the inference phase, due to their increasingly large sizes.
\emph{Early exiting} has proven to be a simple yet effective technique for accelerating inference of LLMs and other deep neural networks.
In this approach, early-exit layers are attached to the original deep neural network, 
which can convert intermediate hidden states into early-exit output.
During inference, the model can adaptively select one early exit to generate the output for each input sample, 
skipping the forward computation of the remaining layers of the network.
Early exiting has found success in 
natural language processing \cite{Graves2016AdaptiveCT,Hou2020,Zhou2020,Schwartz2020,Liu2020,Elbayad2020,Xin2020,Li2021AcceleratingBI,Schuster2021,Xin2021,Xu2023SurveyDynamic,Hu2023SmartBERTAP}, 
computer vision \cite{Panda2015ConditionalDL,Teerapittayanon2016,Kaya2018,Huang2018}, 
and many other areas \cite{Scardapane2020WhySW,Laskaridis2021AdaptiveIT,Han2021DynamicNN,Dai2023ApparateRE}.

This work considers token-wise early exiting for generative LLMs and autoregressive natural language generation
\cite{Schuster2021,DelCorro2023,Bae2023,Varshney2023AcceleratingLI,Gera2023TheBO,chen2023eellm}.
While the majority of prior works in this area have focused on designing early-exit \emph{inference} mechanisms, 
we instead focus on how to \emph{train} an early-exit LLM in the first place.
The standard and straightforward method, adopted in most prior works on early exiting, 
is to jointly train all model parameters (including the network backbone and early-exit layers) from scratch,
by minimizing a weighted sum of training losses from early and final exits.
Recent work has made this approach compatible with massive 3D parallelism,
thereby scaling up early-exit LLMs to sizes as large as any standard LLM that can possibly be trained with state-of-the-art LLM frameworks \cite{Shoeybi2019,Narayanan2021,chen2023eellm}.
The obvious issue with this approach is its excessively high costs and complexity.
Indeed, the massive amount of computational resources required to train an LLM with billions of parameters is simply inaccessible to most members of the community.
Oftentimes in practice, one has access to the weights of an existing pre-trained (and possibly fine-tuned) standard LLM
that might be private or open-source, 
and wonder if it is possible to train an early-exit LLM by leveraging such information rather than from scratch.

\begin{tcolorbox}[
    standard jigsaw,
    size=title,
    opacityback=0,
    left=0.2mm, right=0.2mm, top=0.2mm, bottom=0.2mm]
All these motivate us to \emph{convert an existing generative LLM to an early-exit one}, in a way that 
\begin{itemize}
    \item requires minimum computational resources;
    \item leads to satisfactory inference acceleration; and
    \item preserves the full capability of the original LLM.
\end{itemize}
\end{tcolorbox}

\paragraph{Main contributions.}

\begin{figure}[t]
\centering
\includegraphics[width=1.0\textwidth]{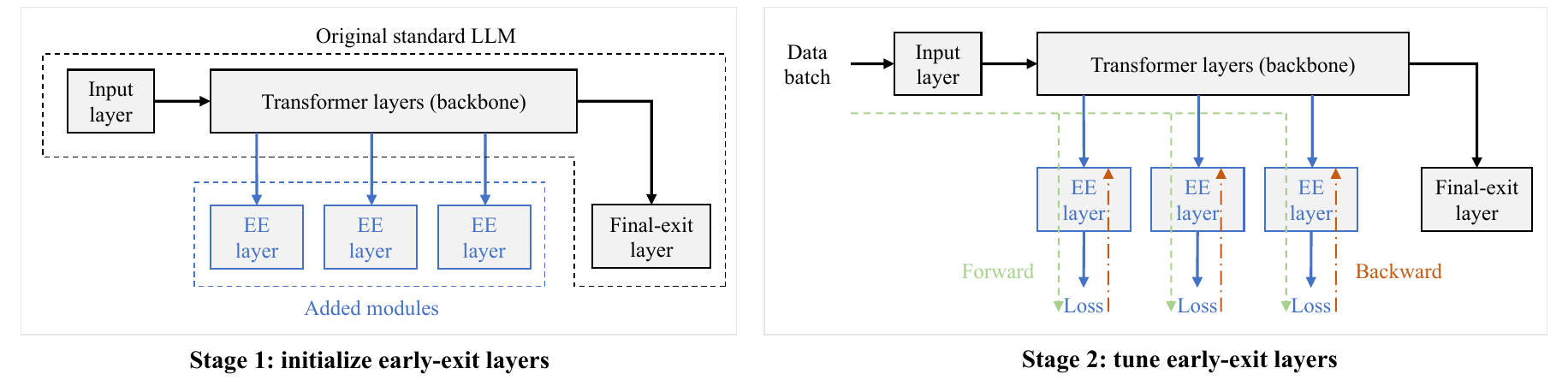}
\caption{An outline of \eetuning, the proposed two-stage procedure that converts a pre-trained standard LLM into a well-trained early-exit LLM.}
\label{fig:ee_tuning_process}
\end{figure}

This work introduces \eetuning, a principled and lightweight approach of transforming a pre-trained (and possibly fine-tuned) LLM into an early-exit one, which satisfies all the above requirements. 
At the core of \eetuning is an intuitive and practical two-stage procedure:
\begin{enumerate}
    \item Take a pre-trained standard LLM as input, and augment its architecture with early-exit layers, whose parameters are initialized properly;
    \item Tune the early-exit layers via backpropagation of certain training losses in a parameter-efficient manner, with modules of the original standard LLM frozen.
\end{enumerate}
See Figure~\ref{fig:ee_tuning_process} for a visualization of our method, 
whose details will be elucidated in Section~\ref{sec:methodology}.
Our implementation is based on the recently proposed \eellm framework \cite{chen2023eellm}, 
complementing the latter with an alternative solution to training early-exit LLMs that is 
both \emph{accessible} and \emph{scalable}, 
thanks to its low computational complexity and full compatibility with 3D parallelism.
In other words, any LLM developer, with access to either one GPU or a cluster with thousands of GPUs, will find \eetuning a useful and practical tool for studying and applying early exiting.
Our implementation also includes support for various configurations and other favorable features, which further makes it more convenient to use.\footnote{Henceforth, we let \eetuning refer to the proposed two-stage method or our implementation of it, depending on the context.}

The efficacy of \eetuning is validated via extensive and systematic experiments
for models with up to 70 billion (70B) parameters, an unprecedented scale for early-exit LLMs.
More specifically, 
a pre-trained LLM can quickly acquire the ability of early exiting via the tuning process with fast and stable convergence,
which takes less than 1/1000 of the GPU hours and training data used in its pre-training stage,
and requires only one or a few GPUs.
Meanwhile, the converted model can achieve $1.2\times$ to $1.6\times$ speedup on various downstream tasks through early exiting while maintaining comparable or even better benchmark scores, or higher speedup if slight degeneration of output quality is acceptable.
We thoroughly investigate the effects of various design choices and 
provide practical guidelines for maximizing the performance of \eetuning.
The source code of \eetuning is available at \url{https://github.com/pan-x-c/EE-LLM}.

\paragraph{Related works.}

The idea of \eetuning, i.e.~augmenting a pre-trained neural network with early-exit layers that are tuned in a parameter-efficient manner, is not completely new.
This strategy has been adopted in some prior works for model architectures tailored to classification tasks, 
e.g.~the encoder-only BERT model \cite{Xin2020,Liu2020,Hu2023SmartBERTAP} or others \cite{Panda2015ConditionalDL,Kaya2018,Bakhtiarnia2021ImprovingTA,Dai2023ApparateRE}.
However, there is no guarantee that results and conclusions from these works can safely transfer to the case of \emph{decoder-only} Transformers tailored to \emph{autoregressive sequence generation}, which is the focus of our work.
Another recent work \cite{Varshney2023AcceleratingLI} proposed to initialize the model parameters of early-exit LLMs with pre-trained standard LLMs, but followed by full-parameter training.
Closest to our setting and training methodology is the recent work \cite{Gera2023TheBO}, 
which investigated generative LLMs of sizes up to 355M, 
and only considered linear exit heads that are randomly initialized.
Moreover, that work proposed to use multiple exits for improving the final output of full-model inference, 
rather than accelerating inference via early exiting.
In contrast, our implementation of the proposed \eetuning method 
(1)~is unified and systematic, with support for a wide range of configurations; 
(2)~is scalable, 
thanks to its full compatibility with 3D parallelism;
and 
(3)~has proven via extensive experiments to return early-exit LLMs that achieve outstanding acceleration during autoregressive inference.

\section{Methodology}
\label{sec:methodology}

This section elaborates on our methodology and implementation of obtaining a well-trained early-exit LLM via \eetuning, 
the two-stage procedure visualized in Figure~\ref{fig:ee_tuning_process}.

\begin{figure*}[t]
\centering
\includegraphics[width=1.0\textwidth]{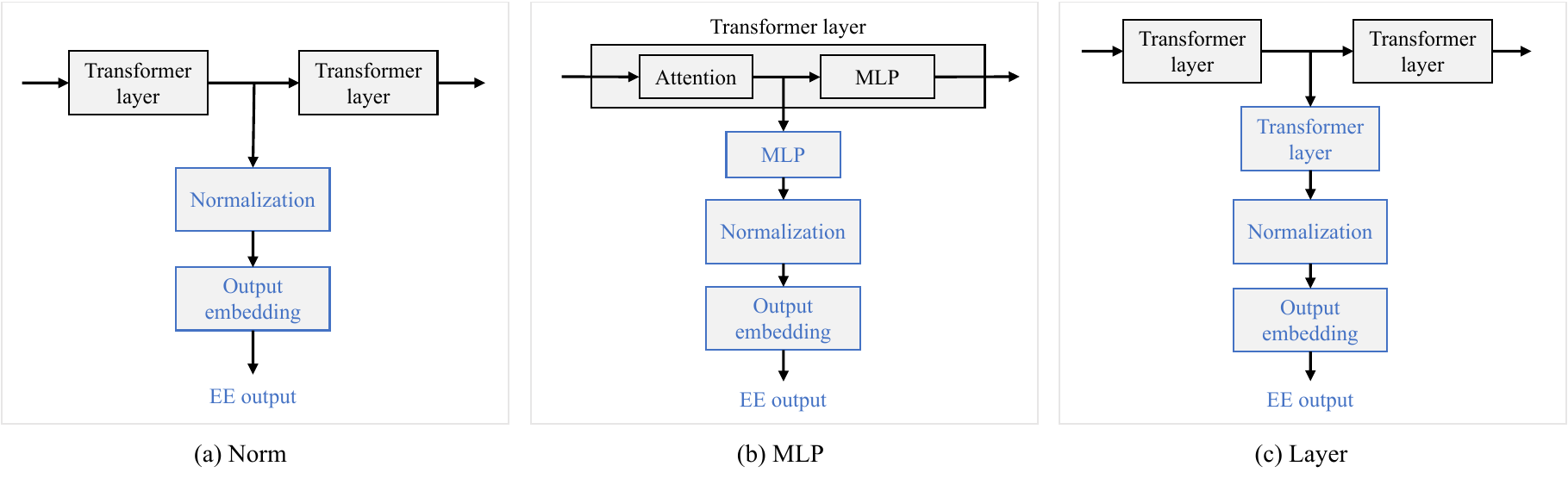}
\caption{A visualization of various early-exit architectures.
Each attention or MLP module follows the residual structure with pre-normalization.
}
\label{fig:EE_structure}
\end{figure*}

\paragraph{Preliminaries.}

Modern LLMs are mostly based on the Transformer architecture \cite{vaswani2017attention}. 
We focus on the decoder-only generative pre-training (GPT) Transformer architecture \cite{Radford2018ImprovingLU,Radford2019LanguageMA}, 
although many techniques presented in this work can be generalized to broader settings.
As visualized in Figure~\ref{fig:ee_tuning_process},
a GPT Transformer is composed of an initial layer for input processing, a stack of Transformer layers as the backbone, and an output layer that converts the final hidden states to logits on the vocabulary, which can be used for generating new tokens.
Each Transformer layer consists of an attention module and a multi-layer perceptron (MLP), with layer normalization \cite{ba2016layernorm} and residual connections \cite{he2016deep} applied in multiple places.
The final output layer includes an optional layer normalization module, followed by a large output embedding matrix.

A GPT Transformer can be trained in an unsupervised manner, by optimizing the language modeling loss on unlabeled corpus.
More specifically, given model parameters  $\btheta$ and a sequence of tokens $\bx = (x_1, x_2, \dots, x_T)$,
the autoregressive language modeling loss, i.e.~negative log-likelihood of next-token prediction, is defined as 
$
\loss(\bx; \btheta) \coloneqq - \log \Pr(\bx; \btheta) 
= - \sum_{t \in [T]} \log \Pr(x_t | x_1, \dots, x_{t-1}; \btheta)
$.

\subsection{Stage 1: initializing early-exit layers}
\label{subsec:method_step_1}

In the first stage of \eetuning, we load a pre-trained standard LLM checkpoint, 
and augment its model architecture by adding early-exit layers to pre-specified locations.
More concretely, we first specify the architecture of early-exit layers, and then initialize their parameters, as explained in the following.

\paragraph{Architectures of early exits.}
In theory, an early-exit layer can be a generic mapping from $\R^h$ to $\R^V$, where $h$ is the hidden size and $V$ is the vocabulary size. 
Our implementation supports the early-exit architectures proposed in prior works \cite{chen2023eellm}, 
which we summarize below.
\begin{itemize}
\item \archembed: This is a minimalistic early-exit architecture with a single linear layer, i.e.~an output embedding matrix of size $h \times V$ that converts hidden states to logits on the vocabulary.
\item \archnorm: A LayerNorm \cite{ba2016layernorm} or RMSNorm \cite{zhang2019root} module, which can stabilize training and improve output quality of the model, is added in front of the \archembed architecture. 
We recommend making the early-exit architecture consistent with the final-exit layer of the original LLM, i.e.~including the normalization module in each early exit if there is one in the final-exit layer.
\item \archmlp: An MLP, with the same structure as the MLPs in the Transformer backbone, is added in front of the \archnorm architecture.
\item \archlayer: A complete Transformer layer, with the same structure as those on the backbone, is added in front of the \archnorm architecture.
\end{itemize}
See Figure~\ref{fig:EE_structure} for a visualization of these architectures.
Generally speaking, a larger number of trainable parameters leads to higher expressivity and adaptivity, but also larger inference latency and potentially higher risk of overfitting during the tuning process, of early-exit layers.
Also note that as a by-product, 
each sub-model between the input of the network and the output of a certain early exit can be regarded as a standard GPT Transformer.

\begin{remark}
Throughout this work, we assume that the output embedding matrices of all exits are untied, unless otherwise specified.
Some recent works on early-exit LLMs choose to tie the output embedding matrices \cite{Schuster2022,Varshney2023AcceleratingLI}, 
and \eetuning can be adapted for this case with some modifications.                 
\end{remark}

\paragraph{Initialization of early exits.}

One natural option for initializing the model parameters is \emph{random} initialization, in the same way as it is done before pre-training a standard LLM from scratch.
To accelerate convergence of \eetuning, we propose a new approach of initializing early-exit layers by \emph{copying} model parameters from certain modules of the original pre-trained LLM.
More specifically,
\begin{itemize}
    \item For \archembed and \archnorm, parameters of the output embedding matrix and normalization module can be copied from the corresponding modules in the final-exit layer of the original LLM;
    \item For \archmlp, the MLP module for an early exit connected to some Transformer layer can be initialized as a duplication of the original MLP within the same Transformer layer;
    \item For \archlayer, the Transformer layer within an exit can be initialized as a duplication of the last Transformer layer of the original LLM.
\end{itemize}

Our proposed approach is primarily inspired by the residual structure \cite{he2016deep} widely adopted in modern LLMs, 
which implies that the output of each Transformer layer is generated by adding a small residual component to the input.\footnote{More rigorously, this is true for Transformers with pre-normalization (the case that we consider in this work), not for post-normalization. Nonetheless, the rationale behind initialization by copying remains mostly correct in the latter case.}
After initialization by copying, the forward pass from the input to the early-exit output is the same as that of the original LLM, except that certain Transformer layers are skipped.
Given the residual structure of the skipped layers, it is reasonable to assume that the early-exit output is still more or less meaningful, which justifies the proposed method of initialization.
Besides accelerating convergence, another potential benefit of initializing by copying is that early-exit layers might inherit the knowledge and abilities that have been learned and embedded in modules of the pre-trained standard LLM.

\subsection{Stage 2: tuning early-exit layers}
\label{subsec:method_step_2}

In the second stage of \eetuning, we tune the early-exit layers via standard backpropagation of training losses from multiple exits, while modules of the original standard LLM are frozen.
By default, we use the autoregressive language modeling loss on open-source datasets of unlabeled corpus, although other options are possible.
Note that, as visualized in Figure~\ref{fig:ee_tuning_process}, this tuning process is equivalent to learning multiple shallow networks independently and in parallel, each of which is an early-exit layer that maps the hidden states at a certain Transformer layer on the backbone to the corresponding early-exit outputs.
Given the relatively small number of trainable parameters, we opt for a \emph{small batch size} in pursuit of fast convergence and good generalization, akin to the standard practice of fine-tuning a small proportion of parameters in a pre-trained deep neural network.

\paragraph{Computational efficiency.}

Our implementation of \eetuning has been well-optimized for maximum computational efficiency.
More specifically, only the minimum amount of necessary computation is executed, including 
(1)~the partial forward pass of the Transformer backbone \emph{up to} the hidden states connected to the last early exit, and
(2)~the forward computation, backward computation, and parameter update for each early-exit layer, without any dependency between early exits.
As for memory usage, we 
(1)~implement partial checkpoint loading, so that only modules of the original LLM in front of the hidden states connected to the last early exit need to be loaded into GPU memory; and
(2)~maintain optimizer states for the early-exit layers only.
With our implementation, \eetuning incurs minor computational overhead compared to a partial forward pass of the original LLM.

\begin{remark}
If one chooses to tie the output embedding matrices from multiple exits, which are untied from the input embedding matrix, 
then \eetuning requires backward computation only for early-exit layers as before.
However, if the output embedding matrices are further tied with the input embedding \cite{Press2017}, 
then a full backward pass through the whole network is needed for gradient calculation, 
which incurs much higher computational costs.
\end{remark}

\begin{figure}
\centering
\includegraphics[width=.5\textwidth]{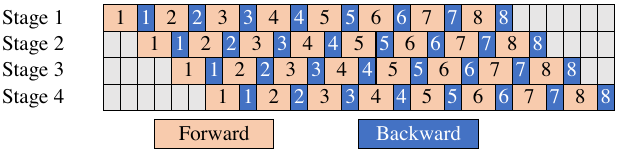}
\caption{One training iteration of our customized pipeline schedule used in \eetuning,
in a setting with 4 pipeline stages and 8 microbatches indexed by numbers in the blocks.
}
\label{fig:EE_tune_pipeline}
\end{figure}

\paragraph{Support for 3D parallelism.}

Built upon prior works \cite{Shoeybi2019,Narayanan2021,chen2023eellm}, our implementation of \eetuning
naturally supports massive 3D parallelism, namely data, tensor, sequence and pipeline parallelism\footnote{With pipeline parallelism, 
a deep neural network is partitioned into multiple pipeline stages along the depth dimension, which are assigned to different computational devices.
In addition, each data batch is divided into multiple microbatches, so that their forward and backward computation on different stages can be pipelined and parallelized via some schedule.}.
For brevity, we refer interested readers to prior works, e.g.~the Megatron-LM series \cite{Shoeybi2019,Narayanan2021,Korthikanti2022,Smith2022}, for details about 3D parallelism, which are omitted here.
It is worth noting that for \eetuning, 
we design and implement a customized pipeline schedule with forward communication only, 
as shown in Figure~\ref{fig:EE_tune_pipeline}.
To see how it boosts training efficiency, recall that for full-parameter training, 
each pipeline stage need to save all intermediate activations of forward computation, 
and wait for the gradients sent backward by the next stage before backward computation can start.
In contrast, things are much easier for \eetuning: 
each early-exit layer within a certain stage can calculate its own loss and execute backward computation independently,
as soon as the forward pass of the Transformer backbone within that stage is completed.
Consequently, there is no need to save intermediate activations on the Transformer backbone, or backward communication for backpropagation.

\subsection{Additional features}
\label{subsec:method_additional}

Some additional features in our implementation of \eetuning are introduced in the following.

\paragraph{Plug-and-play early exits.}

With our implementation,
one can first tune multiple early exits, 
and then decide which one(s) to actually activate for inference, 
in a flexible plug-and-play manner.
Various subsets of the same set of tuned early exits can be chosen for different use cases.
This flexibility of deployment can be beneficial in practice, 
since the best number and locations of early exits are often unknown beforehand, and might depend on the specific use case.

\paragraph{Dynamic token-wise loss weights.}

So far, we have been assuming that each early exit assigns equal weights to all tokens of the training data when defining its training objective.
This, however, causes a deviation from the actual process of early-exit inference, 
where each exit needs to make predictions only for ``easy'' tokens that it has high confidence for, rather than for all tokens.
Some recent works \cite{Tang2023,Regol2023JointlyLearnedEA,Duggal2020ELFAE,Panda2015ConditionalDL,Bakhtiarnia2021ImprovingTA} have proposed to reduce this mismatch by dynamic token-wise loss weights and observed positive outcome,
which motivates us to implement one version of this approach.
More specifically,
during the forward pass of each data batch, each exit records its confidence values for tokens in the batch, which will be used as the weights of tokens for defining the training loss before backward computation.
In this way,
each early exit learns to handle the tokens that it has high confidence in, without being forced to make predictions for ``hard'' tokens that are beyond its capability.

\section{Experiments}

In this section, we validate the efficacy of \eetuning through extensive experiments.
First in Section~\ref{subsec:exp_efficiency}, we demonstrate the efficiency of \eetuning for models of sizes up to 70B, an unprecedented scale of early-exit models in the literature.
Section~\ref{subsec:exp_compare_architectures} investigates the impacts of early-exit architectures on both training losses and downstream performance,
and Section~\ref{subsec:exp_compare_initialization} compares both methods of initializing parameters of early exits.
We further confirm the efficacy of \eetuning for models of various sizes, ranging from 7B to 70B, in Section~\ref{subsec:exp_model_sizes}.
Additional empirical results are deferred to Appendix~\ref{sec:additional_experiments}.

All experiments were conducted with a multi-node GPU cluster, where each node hosts 8 Nvidia A800-80G GPUs.
We explain below the common setup that is adopted throughout our experiments, before presenting the empirical results.

\paragraph{Models.}

For standard LLMs,
we use the open Llama 2-Chat models \cite{Touvron2023llama2} of sizes 7B, 13B and 70B, 
which follow the decoder-only GPT architecture with pre-normalization.
Each model has gone through pre-training with trillions of tokens of training data, and alignment with human preferences via supervised fine-tuning and reinforcement learning from human feedback.
Our early-exit LLMs are constructed by augmenting these \llama models with early-exit layers of various architectures.
For convenience, we denote the two methods of initializing model parameters, which have been introduced in Section~\ref{subsec:method_step_1}, as \initrandom and \initcopy respectively.

\paragraph{Tuning.}

For the tuning process proposed in Section~\ref{subsec:method_step_2},
we set the batch size to a small value of 16, the sequence length to 2048, 
and the number of training iterations to $4 \times 10^{4}$.
Therefore, the total amount of training data for tuning an early-exit LLM is $16 \times 2048 \times 4 \times 10^{4} \approx 1.3$ billion tokens, 
which is almost negligible compared to the number of tokens needed for full-parameter training of a standard or early-exit LLM.
Our training data is randomly sampled from the refined pre-training data provided by Data-Juicer \cite{Chen2023}, unless otherwise specified.
The standard autoregressive language modeling loss, as explained in Section~\ref{sec:methodology}, is used as the training objective in our experiments.
We use the Adam optimizer \cite{Kingma2014AdamAM} with hyperparameters $\beta_1 = 0.9, \beta_2 = 0.95, \epsilon = 10^{-5}$,
together with a learning rate schedule with linear decay, a warmup fraction of $0.01$, a maximum learning rate of $10^{-4}$ and a minimum of $10^{-5}$.

\paragraph{Inference.}

For early-exit inference, we use greedy decoding and a confidence-based exit condition,
i.e.~the model outputs the most likely token as soon as the maximum probability of next-token prediction at some early exit is above a pre-specified threshold.
We utilize the pipeline-based inference mechanism from prior work \cite{chen2023eellm}, which is compatible with KV caching.
Inference efficiency is measured by wall-clock latency.
When the confidence threshold is set to 1, early exits are disabled, so that our early-exit models become equivalent to the original \llama models.
Full-model inference latency in this case is used as the baseline for calculating relative speedup by inference with smaller thresholds.
We conduct downstream evaluation with HELM~\cite{helm} on four tasks, namely CNN/DailyMail \cite{cnndm}, XSUM \cite{xsum}, NarrativeQA \cite{narrativeqa} and MMLU \cite{mmlu}.
For each task, we use a maximum context length of 2048, specify the number of generated tokens, and report 5-shot performance.
See Table~\ref{tab:exp_tasks} for a summary of the tasks.

\begin{table}
\centering
\caption{Tasks for downstream evaluation.}
\label{tab:exp_tasks}
\small
\begin{tabular}{cccc}
\toprule
Task  &  Type  &  Metric  &  Num.~tokens  \\
\midrule
CNN/DailyMail  &  Summarization  &  ROUGE-L  &  128 \\
XSUM            &  Summarization  &  ROUGE-L  &  64  \\
NarrativeQA     &  Reading comprehension  &  F1  &  100 \\
MMLU        &  Language understanding  &  Exact Match  &  1 \\
\bottomrule
\end{tabular}
\end{table}

\subsection{Efficiency of \eetuning}
\label{subsec:exp_efficiency}

\begin{tcolorbox}[
    standard jigsaw,
    size=title,
    opacityback=0,
    left=0.2mm, right=0.2mm, top=0.2mm, bottom=0.2mm]
\textbf{Observation}: \eetuning with Nvidia A800-80G GPUs and without model partitioning can

(1) convert a \llama 13B model within 20 GPU hours, with a peak GPU memory of 20G;

(2) convert a \llama 70B model within 120 GPU hours, with a peak GPU memory of 78G.
\end{tcolorbox}

Our first experiment demonstrates the efficiency of \eetuning.
In this setup, we consider tuning one \archmlp early exit that is added to the 1/4 depth of the \llama model.
We use 4 GPUs for a data parallelism degree of 4, without tensor or pipeline parallelism.
The result is that tuning a 13B model takes about 5 hours and a peak memory usage of 20G.
Such a small memory footprint means that an Nvidia RTX 4090 with 24G memory can also handle this task.
In addition, tuning a 70B model takes about 30 hours and a peak memory usage of 78G, which can fit into a single A800-80G GPU.
Such remarkable efficiency is achievable since with our implementation, only the first 1/4 part of the \llama model checkpoint needs to be loaded into memory, only optimizer states for the early-exit layer need to be maintained, and only the minimum amount of necessary computation is executed.

The curves of training losses at various exits, which are deferred to Section~\ref{subsec:exp_model_sizes} to avoid repetition, confirm the convergence of our tuning process.
Table~\ref{tab:exp_text_with_one_ee} in the appendix shows a few example texts generated by the tuned model,
confirming that \eetuning successfully converts a standard LLM into an early-exit one with accelerated inference.
In the remaining experiments, we examine the performance of early exiting in a more systematic manner.

\subsection{Architectures of early-exit layers}
\label{subsec:exp_compare_architectures}

\begin{tcolorbox}[
    standard jigsaw,
    size=title,
    opacityback=0,
    left=0.2mm, right=0.2mm, top=0.2mm, bottom=0.2mm]
\textbf{Observation}: 
Regarding early-exit architectures,

(1) an early-exit layer with more trainable parameters or located at a deeper position attains a lower training loss;

(2) satisfactory inference speedup can be achieved together with comparable or even boosted scores, 
and \archmlp strikes the best overall balance between speed and quality.





\end{tcolorbox}

This experiment compares the early-exit architectures introduced in Section~\ref{subsec:method_step_1}.
For layer normalization therein, we follow Llama 2 \cite{Touvron2023llama2} and use RMSNorm \cite{zhang2019root}.
We add 8 early exits to the 40-layer 13B \llama model, spaced evenly at the 1/8, 2/8, \dots, 8/8 depth of the Transformer backbone
and initialized by the \initcopy method.
The last early-exit layer, placed side-by-side to the original final-exit layer, is added only for experimental purposes rather than practical usage.

\begin{figure}
\centering
\includegraphics[width=.5\columnwidth]{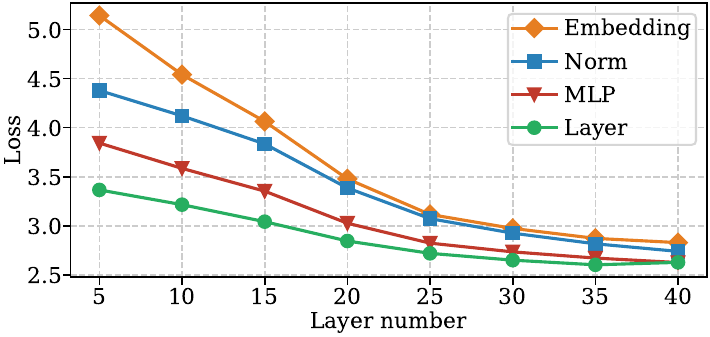}
\caption{Training losses of all early exits at the end of \eetuning for various early-exit architectures.
}
\label{fig:exp_ee_architecture_loss}
\end{figure}

\paragraph{Training losses.}

Figure~\ref{fig:exp_ee_architecture_loss} shows the training loss of each early exit at the end of \eetuning.
We have observed that tuning \archembed early-exit layers, which contain no normalization module, is prone to divergence, or high training losses when it does manage to converge, while tuning exits of the \archnorm architecture can effectively avoid the problem.
Hence, we no longer consider \archembed in our remaining experiments.
Another observation is that in terms of early-exit training losses at the end of tuning, one has $\archnorm > \archmlp > \archlayer$.
This confirms that having more trainable parameters in the early-exit layers leads to lower training losses.
Finally, for each architecture, training losses of early exits at deeper layers are consistently lower than those at earlier layers.
This is unsurprising, since hidden states at deeper layers of a neural network are expected to, on average, contain richer and more usable information about the input, which facilitates better prediction.

\begin{figure}
\centering
\includegraphics[width=.4\columnwidth]{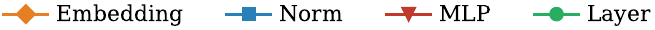}
\includegraphics[width=.25\columnwidth]{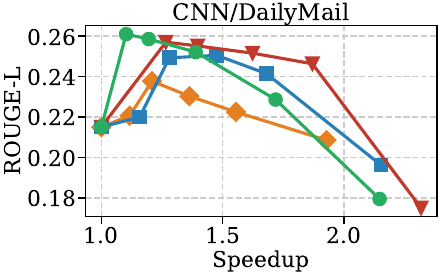}%
\includegraphics[width=.25\columnwidth]{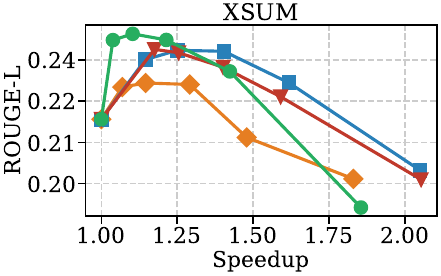}%
\includegraphics[width=.25\columnwidth]{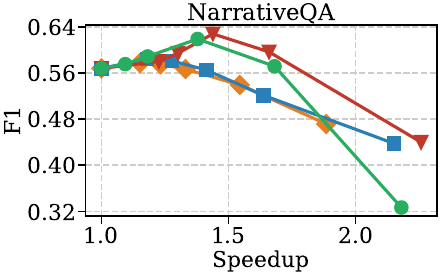}%
\includegraphics[width=.25\columnwidth]{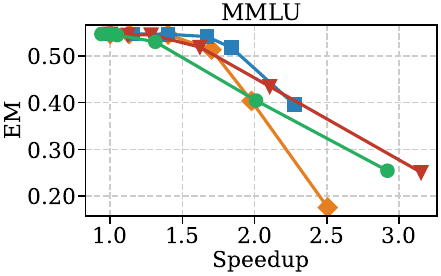}
\caption{Downstream performance of our 13B models with various early-exit architectures. 
Points closer to the top-right corner represent better performance (i.e.~higher speedup and scores).
Markers on each curve correspond to discrete values of the confidence threshold that we use for this experiment.
Speedup increases from left to right as the threshold decreases, taking values in $\{1.0, 0.9, 0.8, 0.6, 0.4, 0.2\}$.
}
\label{fig:exp_ee_architecture_evaluation}
\end{figure}

\paragraph{Inference quality and speedup.}

Figure~\ref{fig:exp_ee_architecture_evaluation} illustrates the downstream performance of early-exit models learned by \eetuning with the aforementioned configurations.
For each model, we activate only three early exits at the 1/4, 2/4, and 3/4 depth, which can be easily done thanks to the plug-and-play feature of our implementation.
On the tasks that we consider, the \archmlp model achieves the best overall performance in terms of scores and speedup, followed by the \archnorm model.
Notably, the \archlayer model exhibits relatively weaker inference speedup, despite achieving the lowest training losses and highest scores in some tasks. 
This is most likely due to the large inference latency of the early-exit layers themselves, as explained in Section~\ref{subsec:method_step_1}.

Careful readers might notice that a higher evaluation score and non-trivial speedup compared to standard full-model inference can be achieved \emph{simultaneously} in some tasks.
While this might seem surprising, similar results have been observed in prior works as well, 
and one reason is that early exiting can help to mitigate \emph{overthinking} \cite{Kaya2018}, 
i.e.~early exits can make the correct prediction while the final exit makes mistakes for certain inputs.
Another possible reason in our case is that, while the original \llama models have been extensively fine-tuned for alignment with human preferences, the additional early-exit layers have only been tuned with the language modeling objective on pre-training data, which incurs less \emph{alignment tax} and hence a slight advantage over the full-model output in certain benchmarks.
Finally, regarding the risk of data leakage during \eetuning, 
we recall that our training data is a small subset of 1.3B tokens randomly sampled from commonly used open-source datasets for LLM pre-training \cite{Chen2023}.
Therefore, it is reasonable to assume that the early-exit layers have seen very few, if any, samples of the evaluation tasks during \eetuning.

\subsection{Initialization of early-exit layers}
\label{subsec:exp_compare_initialization}

\begin{figure}
\centering
\includegraphics[width=.5\columnwidth]{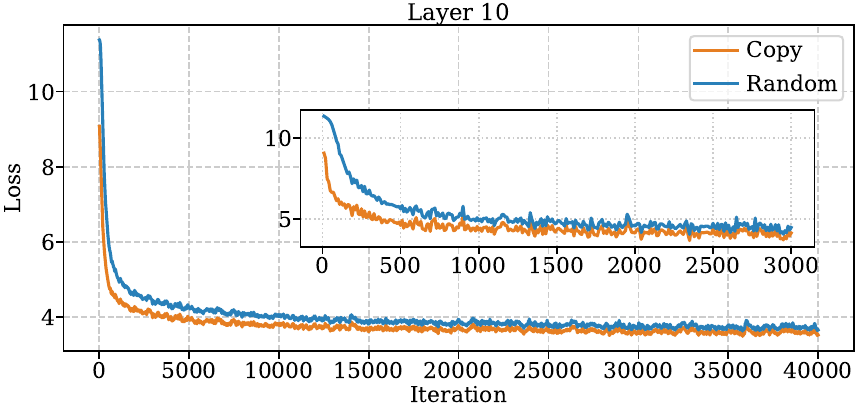}%
\includegraphics[width=.5\columnwidth]{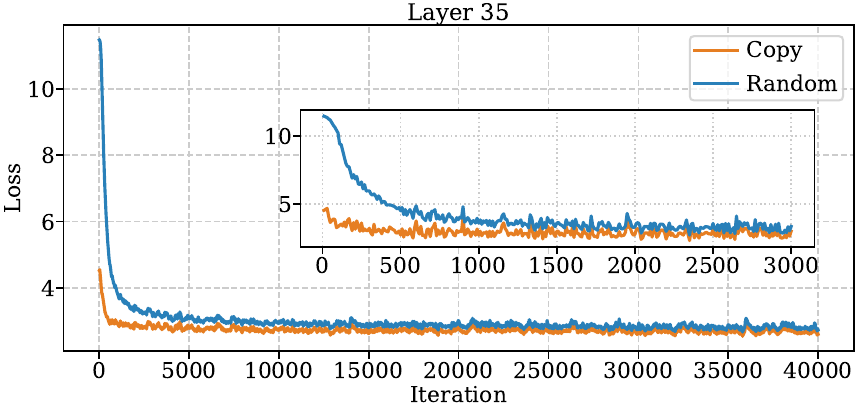}
\caption{Loss curves at two exits, initialized by either \initcopy or \initrandom. 
See Figure~\ref{fig:exp_ee_init_loss_complete} for the complete version with loss curves for 8 exits.
}
\label{fig:exp_ee_init_loss}
\end{figure}

\begin{figure}
\centering
\includegraphics[width=.2\columnwidth]{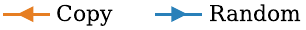}
\includegraphics[width=.25\columnwidth]{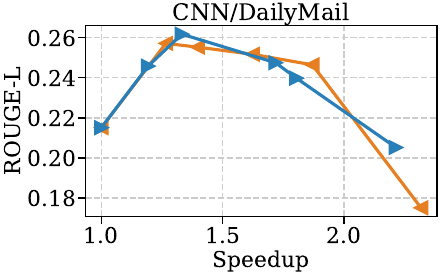}%
\includegraphics[width=.25\columnwidth]{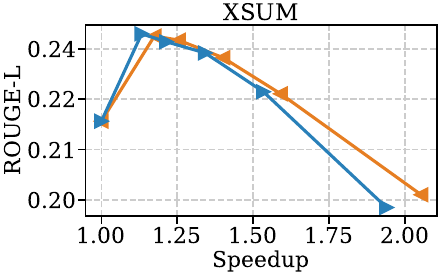}%
\includegraphics[width=.25\columnwidth]{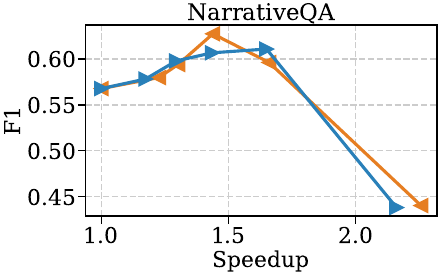}%
\includegraphics[width=.25\columnwidth]{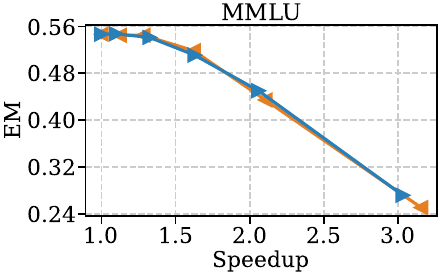}
\caption{Downstream performance of models with early-exit layers initialized by either \initcopy or \initrandom.}
\label{fig:exp_ee_init_evaluation}
\end{figure}

\begin{tcolorbox}[
    standard jigsaw,
    size=title,
    opacityback=0,
    left=0.2mm, right=0.2mm, top=0.2mm, bottom=0.2mm]
\textbf{Observation}: 
\initcopy incurs faster convergence and lower losses compared to \initrandom, while the final inference quality and speedup are similar in both cases.
\end{tcolorbox}

This experiment compares two methods introduced in Section~\ref{subsec:method_step_1}, 
namely \initcopy and \initrandom, of initializing the model parameters of early-exit layers.
We follow the same setup as in Section~\ref{subsec:exp_compare_architectures},
and add one case where \archmlp early-exit layers are initialized with normal random variables.



Figure~\ref{fig:exp_ee_init_loss} demonstrates the training loss curve at each early exit for either method of initialization.
We observe that initialization by \initcopy leads to much lower losses and faster convergence during the early stage of tuning (especially for exits at deeper layers), although training losses in the \initrandom cases will eventually catch up, up to marginal gaps.
While we fix the number of tuning iterations throughout our experiments for simplicity, 
such results suggest that with \initcopy, one might further improve training efficiency by terminating the tuning process earlier, as soon as losses saturate.
Figure~\ref{fig:exp_ee_init_evaluation} further shows that there is no significant difference in downstream performance between models initialized by \initcopy or \initrandom.

\subsection{\eetuning for models of various sizes}
\label{subsec:exp_model_sizes}

\begin{tcolorbox}[
    standard jigsaw,
    size=title,
    opacityback=0,
    left=0.2mm, right=0.2mm, top=0.2mm, bottom=0.2mm]
\textbf{Observation}: 
\eetuning converges smoothly and attains $1.2\times$ to $1.6\times$ inference speedup without sacrificing output quality for models of different sizes,
and larger models achieve better speedup in general (except for MMLU).
\end{tcolorbox}



\begin{figure}
\centering
\begin{subfigure}{.5\columnwidth}
    \centering
    \includegraphics[width=\textwidth]{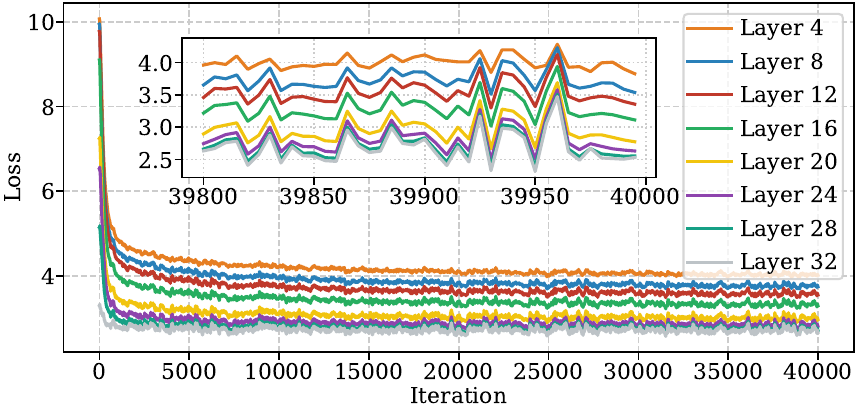}
    \caption{7B}    
\end{subfigure}
\begin{subfigure}{.5\columnwidth}
    \centering
    \includegraphics[width=\textwidth]{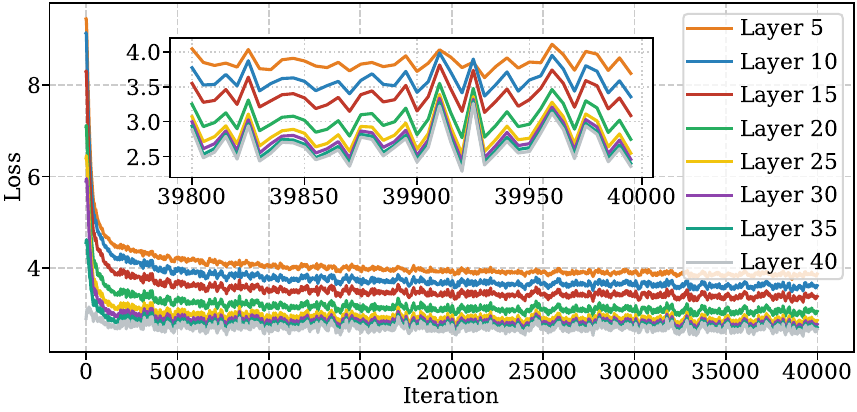}
    \caption{13B}    
\end{subfigure}
\begin{subfigure}{.5\columnwidth}
    \centering
    \includegraphics[width=\textwidth]{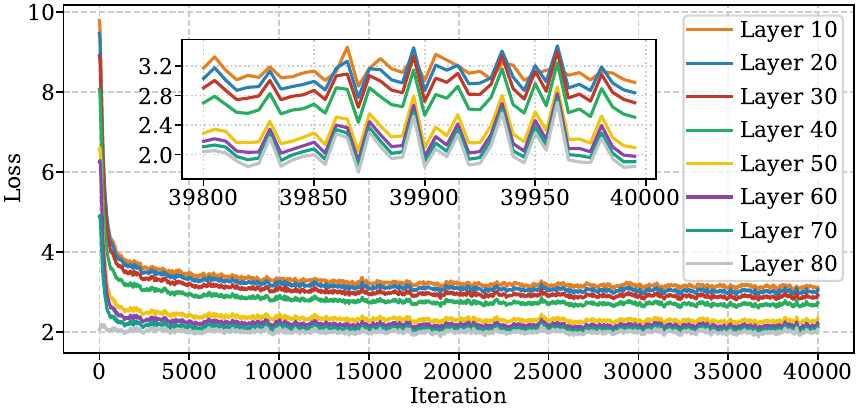}
    \caption{70B}    
\end{subfigure}
\caption{Training loss curves of \eetuning for models of sizes ranging from 7B to 70B.
}
\label{fig:exp_various_size_loss}
\end{figure}

\begin{figure}
\centering
\includegraphics[width=.25\columnwidth]{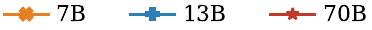}
\includegraphics[width=.25\columnwidth]{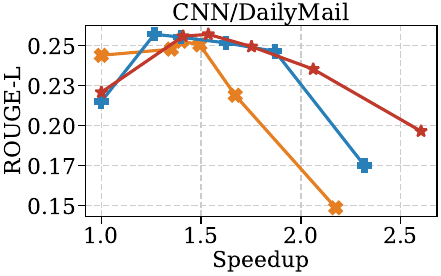}%
\includegraphics[width=.25\columnwidth]{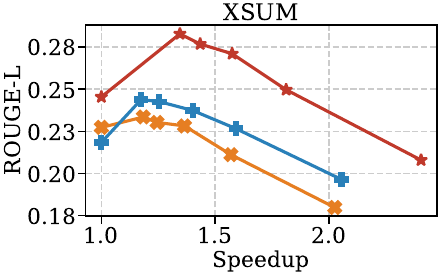}%
\includegraphics[width=.25\columnwidth]{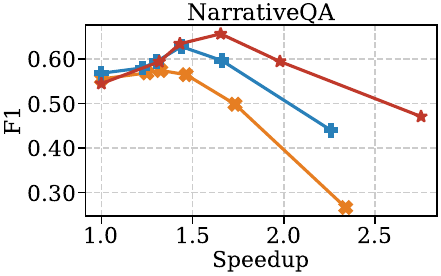}%
\includegraphics[width=.25\columnwidth]{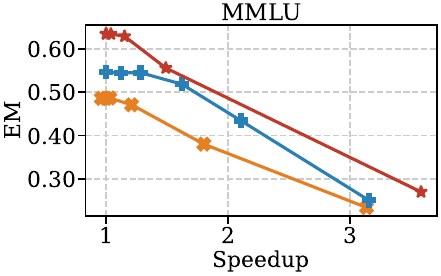}
\caption{Downstream performance of early-exit models of various sizes.
For each model, early exits at the 1/4, 2/4, and 3/4 depth are activated.
Speedup increases from left to right as the confidence threshold decreases, taking values in $\{1.0, 0.9, 0.8, 0.6, 0.4, 0.2\}$.
}
\label{fig:exp_various_size_evaluation}
\end{figure}

This experiment validates the efficacy of \eetuning for LLMs of various sizes.
While previous experiments focus on 13B models, here we consider \llama models of sizes 7B, 13B or 70B.
For each model, we add 8 early exits with the \archmlp architecture,
which are spaced evenly on the Transformer backbone, 
and initialized by the \initcopy method.


Figure~\ref{fig:exp_various_size_loss} shows the training loss curves of all exits for each model, 
confirming the convergence of \eetuning for models of various sizes.
Unsurprising, larger models achieve lower training losses, 
and exits at deeper layers achieve lower losses within each model.
Figure~\ref{fig:exp_various_size_evaluation} further confirms that each model, 
with three early exits at the 1/4, 2/4 and 3/4 depth activated,
achieves early-exit speedup with comparable or sometimes higher scores than full-model inference in the downstream tasks that we consider.

\subsection{Additional experiments}

We have explored other aspects of \eetuning, including
(1)~different sources of training data for the tuning process;
(2)~choosing the best subset of tuned early exits to activate for inference;
(3)~downstream performance of sub-models 
induced by early exits;
(4)~differences between tuning with static or token-wise dynamic loss weights proposed in Section~\ref{subsec:method_additional}; and
(5)~potential benefits of continued pre-training with full-model parameter updating after \eetuning is completed.
The results can be found in Appendix~\ref{sec:additional_experiments}.

\section{Limitations and future work}
\label{sec:limitations}

\paragraph{\eetuning vs.~joint training.}

In \eetuning, early-exit layers are tuned while modules of the original standard LLM are frozen.
As a result, the tuning process is highly efficient, and has no impact on the full-model output.
The obvious disadvantage is that expressivity and adaptivity of early exits are limited, 
since the Transformer backbone was originally trained to produce intermediate hidden states that are useful for generating the full-model output, rather than early-exit output.
In other words, the capability of early exits is inevitably constrained by the limited number of trainable parameters.
When sufficient computational resources are available,
a natural strategy to further improve the tuned early-exit model is joint learning of both network backbone and early exits, 
via full-parameter continued pre-training (CPT) or parameter-efficient fine-tuning like LoRA \cite{Hu2022}.
Indeed, our preliminary experimental result in Appendix~\ref{subsec:exp_cpt} confirms that early-exit losses continue to decay smoothly during CPT.
It would be interesting to see how training efficiency and downstream performance of \eetuning $+$ CPT compare to those of pre-training from scratch, 
as well as understand how the learning dynamics in both cases differ.


\paragraph{Other training objective.}

One potential improvement for the \eetuning method is to use more general training objective, 
beyond the autoregressive language modeling loss on pre-training data.
For example, given that the original standard LLM has been well pre-trained and fine-tuned,
it can serve as the teacher in knowledge distillation \cite{Hinton2015DistillingTK}, 
and supervise the training of early-exit layers using its own output logits as soft labels.
This approach, sometimes called self-distillation \cite{Liu2020,Zhang2021SelfDistillationTE}, has been adopted in some prior works for training early-exit models from scratch, while we find it particularly appropriate for the setting of \eetuning.
One might even consider using texts generated by the original LLM as training data for \eetuning, without relying on external pre-training data that is noisy and potentially contains harmful or undesirable contents.
With such modifications, the tuned early exits might better inherit the knowledge and abilities of the original LLM.

\paragraph{Limited configurations for experiments.}

While we have tried to make our experiments as extensive as we can afford,
there are still many factors that were fixed throughout.
For example, we have only tried out \llama models for initializing our early-exit LLMs, 
and used one inference mechanism (greedy decoding and confidence-based exit condition) for all downstream evaluation.
In addition, hyperparameters for training are the same for models of various sizes,
although it might be more reasonable to use larger batch size and total number of tokens for tuning a 70B model than for a 7B model.
Consequently, one should be cautious about extrapolating observations and conclusions from our experiments to broader settings.
Output quality and speed during early-exit inference might be further improved by better choices of training configurations or inference/decoding mechanisms.

\paragraph{Lack of fine-tuning for alignment.}

Our early-exit models were initialized with \llama models, which have been fine-tuned for alignment with human preferences \cite{Touvron2023llama2}.
The early-exit layers, however, were tuned only with pre-training data and language modeling losses.
We conjecture that the tuned early-exit models preserve most of the alignment properties of the original \llama models, due to the relatively small number of model parameters in early-exit layers.
With that said, we have not empirically evaluated relevant metrics, such as helpfulness and safety, of our early-exit models.
Caution must be taken before deploying these models, and an extra fine-tuning stage for better alignment might be helpful.

\section{Conclusions}

This work has provided a unified and systematic study of \eetuning, 
a lightweight and economical approach to converting any existing LLM into an early-exit LLM in a parameter-efficient manner.
Our implementation of \eetuning is well optimized for maximum computational efficiency,
and also highly scalable thanks to its compatibility with massive 3D parallelism.
Results of extensive experiments have validated that,
with negligible training costs compared to full-parameter training,
\eetuning successfully returns early-exit LLMs that achieve outstanding speedup with minor or no degeneration of output quality during the inference phase.
It is our hope that this work will make early-exit LLMs more accessible to the community.

\begin{appendices}

\section{Additional experiments}
\label{sec:additional_experiments}

This section includes additional experiments and empirical results for \eetuning.

\subsection{Training data for \eetuning}

This experiment explores the impacts of the training data used for \eetuning.
Similar to previous experiments, we consider a 13B model with 8 \archmlp early exits
evenly spaced on the Transformer backbone and initialized by the \initcopy method.
The only difference is that, 
instead of the \emph{pre-training} data used in previous experiments,
here we use the \emph{instruction fine-tuning} (IFT) data\footnote{\url{https://huggingface.co/datasets/datajuicer/alpaca-cot-en-refined-by-data-juicer}} provided by Data-Juicer \cite{Chen2023}, 
which is a refined subset of the Alpaca-CoT dataset \cite{si2023empirical},
for the tuning process.
Generally speaking, the pre-training data is more diverse, 
while the IFT data is cleaner and more structured, organized in a query-answer format.

Figure~\ref{fig:exp_data_loss} shows the training losses at different early exits, both at the beginning and at the end of the tuning process.
Compared with \pt data, using \ift data incurs higher losses initially, but then lower losses after tuning is completed.
This is possible because the early exits quickly adapt to the format and style of the \ift data during tuning; 
thereafter, lower losses can be achieved since the \ift data is cleaner, more structured, and less diverse.


Figure~\ref{fig:exp_ee_architecture_evaluation_ift} illustrates the downstream performance of models with various early-exit architectures.
It is similar to Figure~\ref{fig:exp_ee_architecture_evaluation}, except that tuning is done with \ift rather than \pt data.
We further take the curves from both figures corresponding to the \archmlp architecture,
and compare them side by side in Figure~\ref{fig:exp_data_evaluation},
which shows that the model tuned with \pt data indeed outperforms the one tuned with \ift data.
Table~\ref{tab:exp_data_example_text} in the appendix
demonstrates some example texts generated by both models.
We notice that the model tuned with \ift data sometimes uses undesirable formats in its output instead of strictly following the formats of in-context examples, 
which could possibly be caused by overfitting to \ift data to some extent.
These results suggest that it might be more reasonable to first use \pt data for \eetuning, 
which allows the added early exits to acquire general language abilities;
thereafter, \ift data can be used for further fine-tuning.

\begin{figure}
\centering
\includegraphics[width=.3\textwidth]{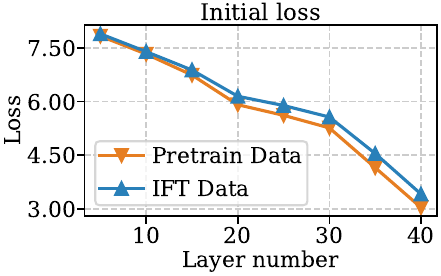}%
\includegraphics[width=.3\textwidth]{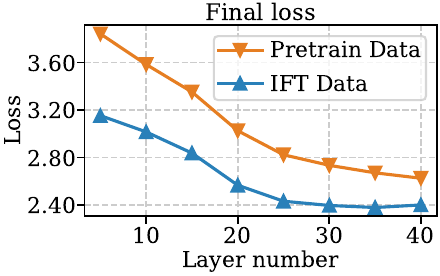}
\caption{Training losses of \archmlp early exits at the beginning (left) or end (right) of the tuning process with either \pt or \ift data.
}
\label{fig:exp_data_loss}
\end{figure}

\begin{figure}
\centering
\includegraphics[width=.5\textwidth]{figs/eval/structure/legend.pdf}
\includegraphics[width=.25\textwidth]{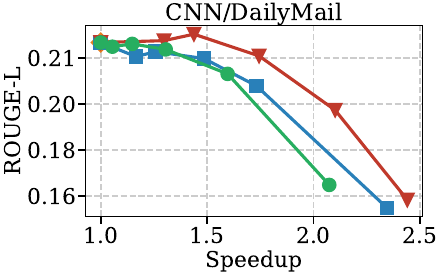}%
\includegraphics[width=.25\textwidth]{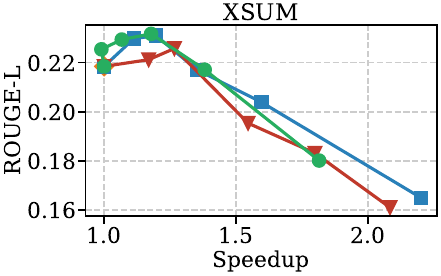}%
\includegraphics[width=.25\textwidth]{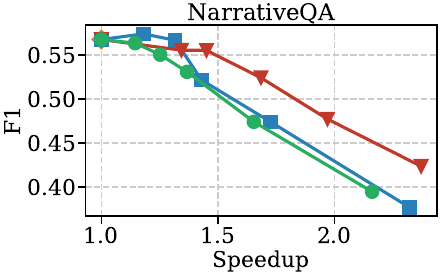}%
\includegraphics[width=.25\textwidth]{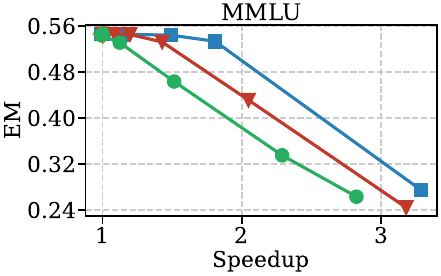}
\caption{Downstream performance of our 13B models with various early-exit architectures, tuned with \ift data.
For each curve, the confidence threshold decreases from left to right, taking values in $\{1.0, 0.9, 0.8, 0.6, 0.4, 0.2\}$.
}
\label{fig:exp_ee_architecture_evaluation_ift}
\end{figure}

\begin{figure}
\centering
\includegraphics[width=.25\textwidth]{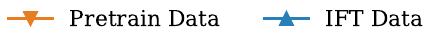}
\includegraphics[width=.25\textwidth]{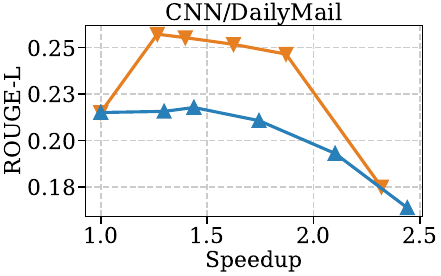}%
\includegraphics[width=.25\textwidth]{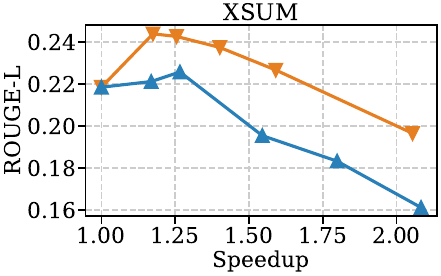}%
\includegraphics[width=.25\textwidth]{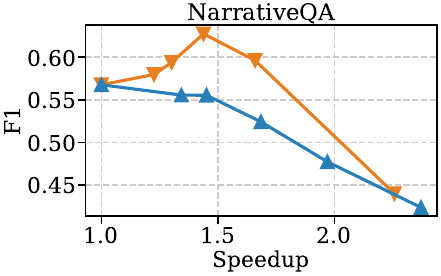}%
\includegraphics[width=.25\textwidth]{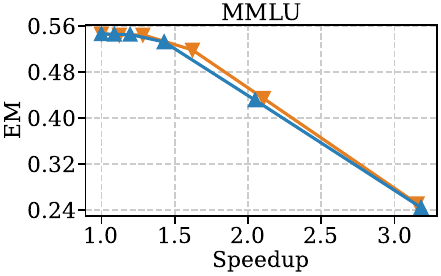}
\caption{A side-by-side comparison between the downstream performance of early-exit LLMs tuned with \pt or \ift data.
The results here are duplication of the curves in Figures~\ref{fig:exp_ee_architecture_evaluation} and~\ref{fig:exp_ee_architecture_evaluation_ift} corresponding to the \archmlp architecture.}
\label{fig:exp_data_evaluation}
\end{figure}

\subsection{Supplementary results for Section~\ref{subsec:exp_model_sizes}}

Below are additional empirical results for our previous experiment in Section~\ref{subsec:exp_model_sizes} 
that validates the efficacy of \eetuning for models of various sizes.

\paragraph{Choosing which early exit(s) to activate.}

Given multiple tuned early exits, one might wonder which subset should be activated for the best inference performance.
For a preliminary exploration, we restrict ourselves to activating only one single early exit.
Scores and speedup in downstream tasks for each option can be found in Figure~\ref{fig:exp_one_EE_evaluation}.
The key trade-off here is that, early exits at deeper layers generally have higher capability, but also larger inference latency.
Based on our empirical results, the sweet spot seems to vary case by case,
and thus we recommend choosing the subset of activated early exits via standard hyperparameter optimization on a validation set before deployment.


\begin{figure*}
\centering
\includegraphics[width=.25\textwidth]{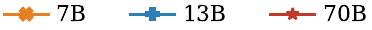}
\includegraphics[width=.25\textwidth]{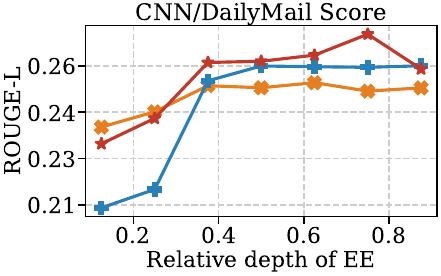}%
\includegraphics[width=.25\textwidth]{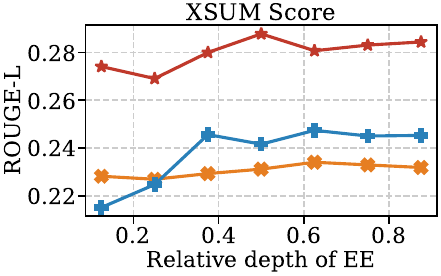}%
\includegraphics[width=.25\textwidth]{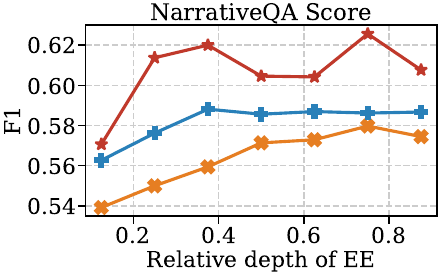}%
\includegraphics[width=.25\textwidth]{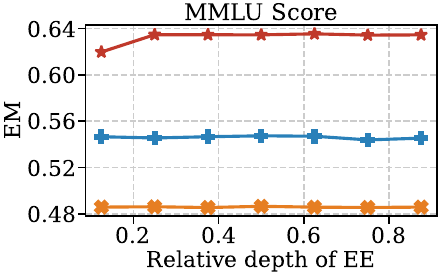}
\includegraphics[width=.25\textwidth]{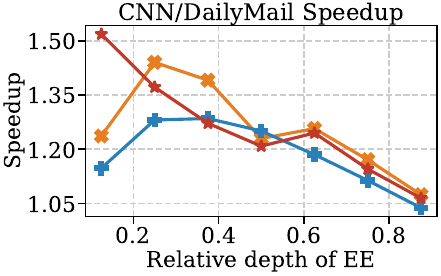}%
\includegraphics[width=.25\textwidth]{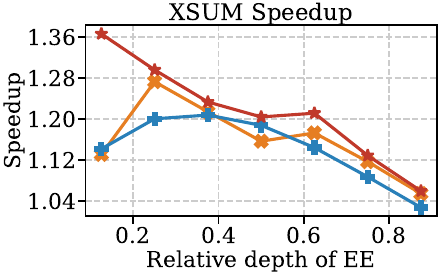}%
\includegraphics[width=.25\textwidth]{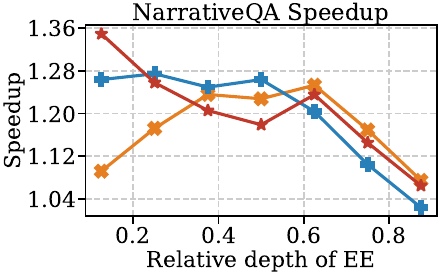}%
\includegraphics[width=.25\textwidth]{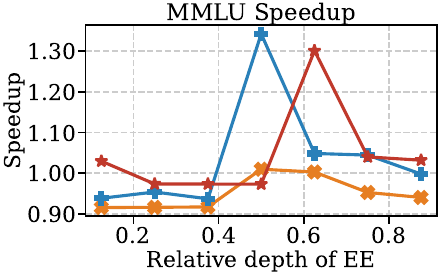}
\caption{Scores (top) and speedup (bottom) in downstream tasks, 
with different choices of activating one single early exit.
The confidence threshold is fixed at 0.8.
}
\label{fig:exp_one_EE_evaluation}
\end{figure*}

\paragraph{By-products: sub-models induced by early exits.}

Recall from Section~\ref{subsec:method_step_1} that the sub-model induced by each early exit,
which includes the modules covered by a forward pass from the beginning of the network to the output of the early exit,
can be regarded as a standard Transformer model with fewer Transformer layers than the original full model.
Such sub-models might be deployed for standard LLM inference,
which can also be regarded as a simplified mechanism of early-exit inference, 
namely using the same pre-specified early exit for generating all tokens of a sequence.
Indeed, empirical results in Figure~\ref{fig:exp_submodel_evaluation} confirm that these sub-models,
especially those corresponding to early exits at deep layers,
performs reasonably well in downstream tasks.

\begin{figure*}
\centering
\includegraphics[width=.25\textwidth]{figs/eval/size/legend.pdf} 
\includegraphics[width=.25\textwidth]{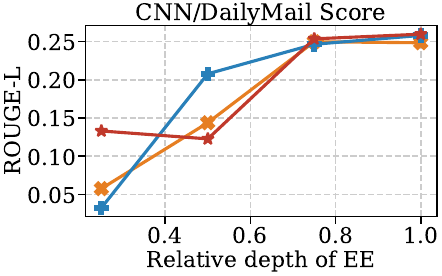}%
\includegraphics[width=.25\textwidth]{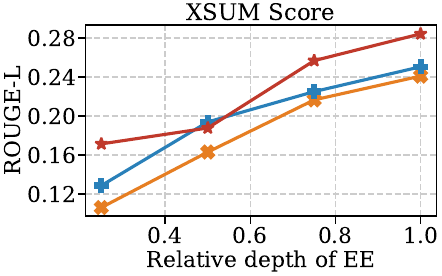}%
\includegraphics[width=.25\textwidth]{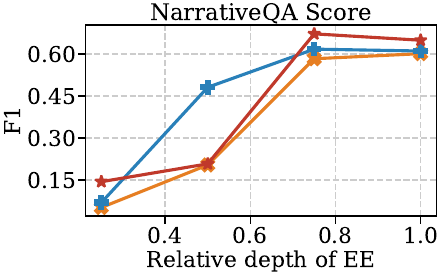}%
\includegraphics[width=.25\textwidth]{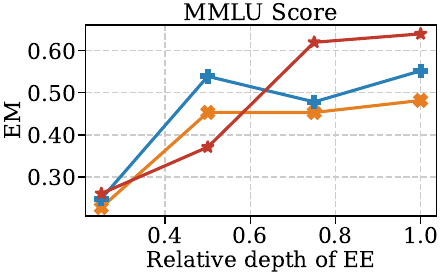}
\caption{Downstream performance of standard inference with sub-models induced by early exits.}
\label{fig:exp_submodel_evaluation}
\end{figure*}

\subsection{Dynamic token-wise loss weighting}

This section provides a preliminary exploration of using dynamic token-wise loss weights in \eetuning, 
as explained in Section~\ref{subsec:method_additional}.
More specifically, let us denote the model parameters at the current training iteration as $\btheta$;
in addition, consider a sequence of tokens $\bx = [x_1, x_2, \dots, x_T]$ in the data batch,
and let $\bc = [c_1, c_2, \dots, c_T] \in [0, 1]^{T}$ be the confidence values (i.e.~the maximum probabilities of next-token prediction) calculated at a certain early exit during the forward pass, which are detached from the computational graph and thus regarded as constants.
Then, the training loss for this sequence at that early exit is defined as
$$
\loss(\bx; \btheta) \coloneqq - \sum_{t \in [T]} c_t \cdot \log \Pr(x_t | x_1, \dots, x_{t-1}; \btheta).
$$
In other words, the negative log-likelihood for each token is weighted by the corresponding confidence value.
Ideally, this method encourages each early-exit layer to learn to predict tokens that are within its capability.

To verify the effectiveness of this method, we conduct an experiment based on the intermediate checkpoint of the 13B \archmlp model in Section~\ref{subsec:exp_compare_architectures}, which has been tuned on half of the training data.
We continue the tuning process on the remaining half of the data with dynamic token-wise loss weighting enabled, while other hyperparameters remain unchanged.
Figure~\ref{fig:exp_dynamic_weight_evaluation} compares the downstream performance of the model obtained by the above method and the original model that was tuned using constant loss weights throughout.
Unfortunately, we see no obvious difference between them.
This may be caused by  
(i)~the short period of training with dynamic weighting; 
(ii)~the small learning rate in the second half of training; or
(iii)~the small number of trainable parameters.
The real reason behind this, and the right way to achieve gains from dynamic weighting, 
need to be further investigated in future works.

\begin{figure}
\centering
\includegraphics[width=.30\textwidth]{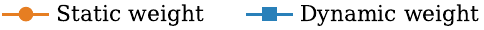}
\includegraphics[width=.25\textwidth]{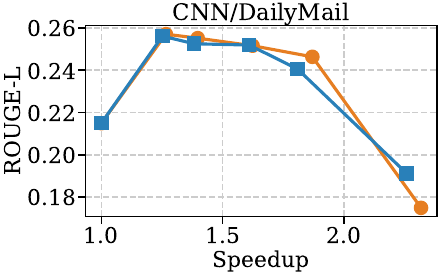}%
\includegraphics[width=.25\textwidth]{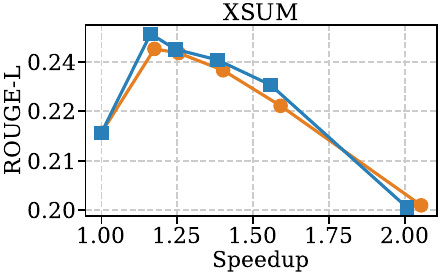}%
\includegraphics[width=.25\textwidth]{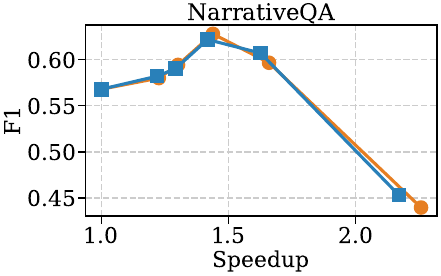}%
\includegraphics[width=.25\textwidth]{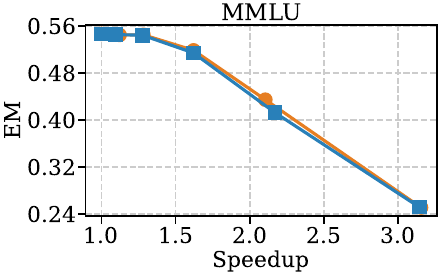}
\caption{Downstream performance of models tuned with constant or dynamic loss weights.}
\label{fig:exp_dynamic_weight_evaluation}
\end{figure}

\subsection{Continued pre-training (CPT) after \eetuning}
\label{subsec:exp_cpt}

To understand whether full-parameter CPT can further improve the early-exit LLMs obtained via \eetuning,
we perform CPT for our 7B early-exit model using the scripts for pre-training provided by \eellm \cite{chen2023eellm}, except that a smaller learning rate is used.
Training loss curves for two early exits and the final exit can be found in Figure~\ref{fig:exp_cpt}.
Interestingly, we observe that the early-exit losses continue to decay smoothly,
while the final-exit loss drops quickly during the first few iterations, and then remains constant.
Such results suggest that CPT might further improve the early exits, without negatively impacting the full-model output.

\begin{figure}
\centering
\includegraphics[width=.5\textwidth]{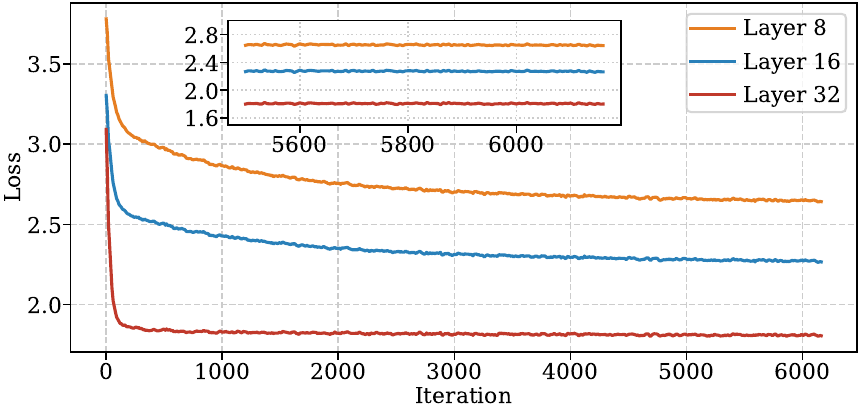}
\caption{Convergence of training losses during CPT after \eetuning is completed.}
\label{fig:exp_cpt}
\end{figure}


\clearpage

\begin{figure*}
\centering
\includegraphics[width=.5\textwidth]{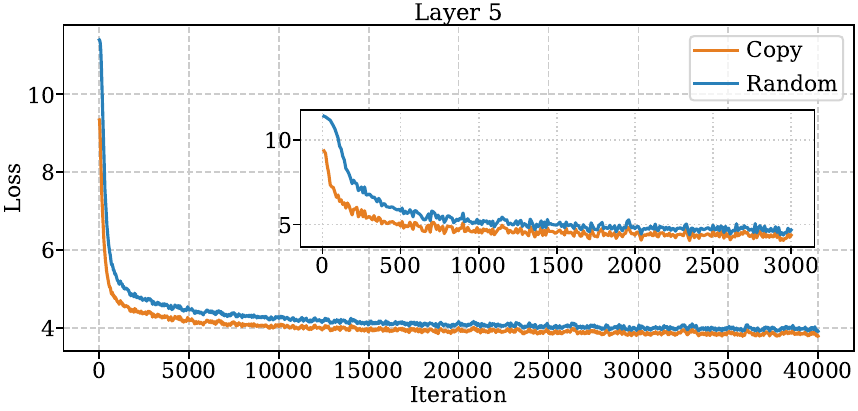}%
\includegraphics[width=.5\textwidth]{figs/loss/init/10-pt.pdf}
\includegraphics[width=.5\textwidth]{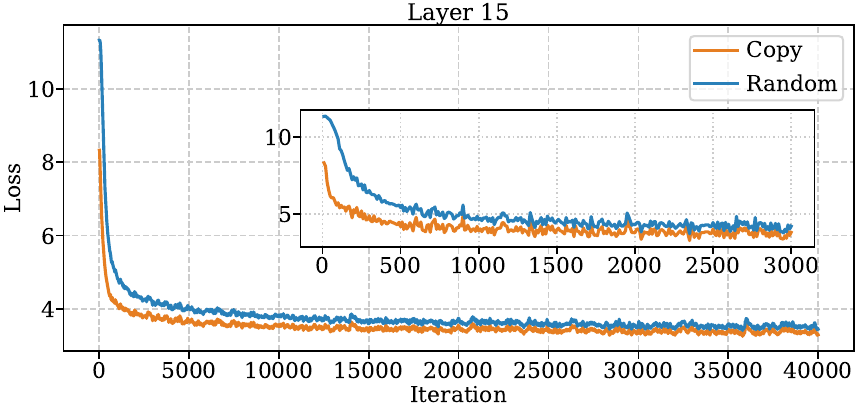}%
\includegraphics[width=.5\textwidth]{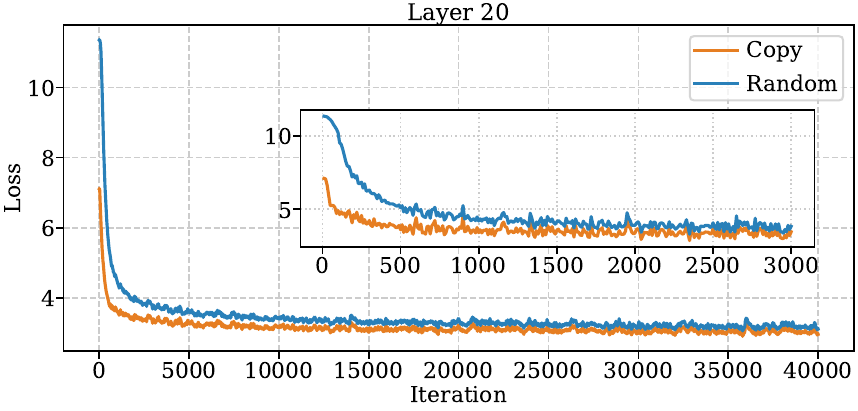}
\includegraphics[width=.5\textwidth]{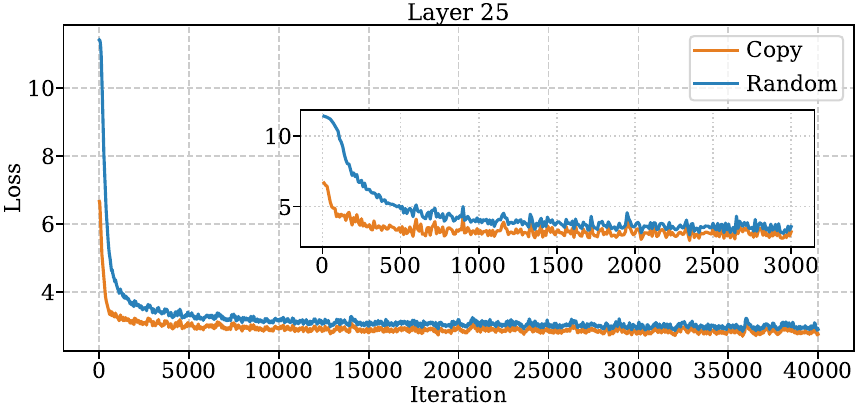}%
\includegraphics[width=.5\textwidth]{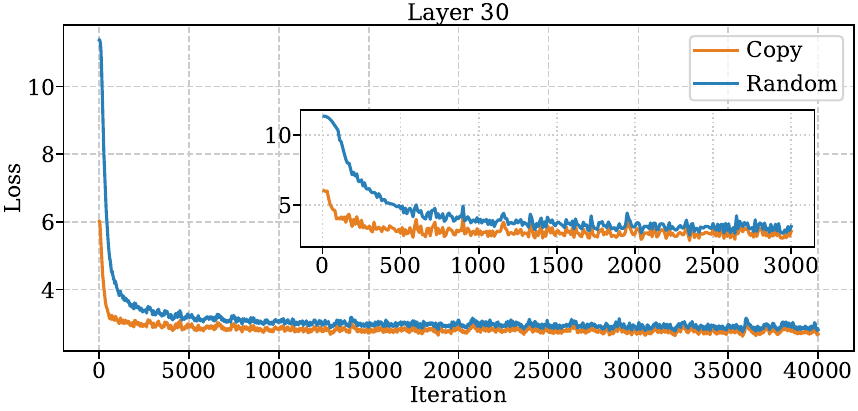}
\includegraphics[width=.5\textwidth]{figs/loss/init/35-pt.pdf}%
\includegraphics[width=.5\textwidth]{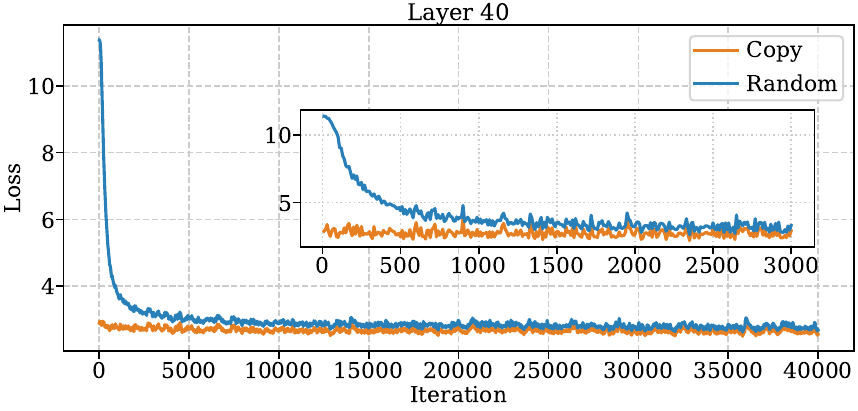}
\caption{Training loss curves at each exit, initialized by either \initcopy or \initrandom. 
}
\label{fig:exp_ee_init_loss_complete}
\end{figure*}

\clearpage




\begin{table}
\centering
\caption{Example texts (and the corresponding inference latencies) generated by a 13B or 70B early-exit LLM having one \archmlp early exit at the 1/4 depth of its Transformer backbone.
Differences from the texts generated by full-model inference are highlighted.
}
\label{tab:exp_text_with_one_ee}
\footnotesize
\begin{tabular}{p{.15\textwidth}p{.78\textwidth}}
    \toprule
    Prompt & Artificial General Intelligence is \\
    \midrule
    13B\newline Full model\newline Time: 4.8 s & the holy grail of AI research. It is the ability of an AI system to perform any intellectual task that a human can, and to do so at a level of performance that is indistinguishable from a human. AGI would be a system that can learn, reason, and self-improve at a pace that is exponentially faster than human intelligence.\newline There are many challenges to achieving AGI, including:\newline 1. Lack of \\
    \midrule[.01pt]
    13B\newline Threshold = 0.8\newline Time: 4.2 s & the holy grail of AI research. It is the ability {\color{purple}to create a machine} that can perform any intellectual task that a human can, and {\color{purple}potentially even surpass human-level performance. However, achieving AGI is a daunting task, and many experts believe that it is still decades away. In this article, we will explore the current state of AGI research, the challenges that researchers face, and the potential implications of achieving AGI.} \\
    \midrule[.01pt]
    13B\newline Threshold = 0.6\newline Time: 3.9 s & the holy grail of AI research. It is the ability {\color{purple}to create a machine that can} perform any intellectual task that a human can, and {\color{purple}potentially even surpass human-level performance. However, achieving AGI is a daunting task, and many experts believe that it is still decades away.\newline One of the main challenges in achieving AGI is the problem of "explainability." As AI systems become more complex and sophisticated, it} \\
    \midrule
    70B\newline Full model\newline Time: 23.6 s & a hypothetical AI that possesses the ability to understand, learn, and apply knowledge across a wide range of tasks, much like human intelligence. While we have made significant progress in developing AI systems that can perform specific tasks, such as image recognition, natural language processing, and autonomous driving, we are still far from achieving true AGI.\newline One of the main challenges in developing AGI is the lack of understanding of human intelligence. While we have made significant progress in \\
    \midrule[.01pt]
    70B\newline Threshold = 0.8\newline Time: 19.7 s & a hypothetical AI that possesses the ability to understand, learn, and apply knowledge across a wide range of tasks, much like human intelligence. While we have made significant progress in developing AI systems that can perform specific tasks, such as image recognition, natural language processing, and autonomous driving, we are still far from achieving true AGI.\newline One of the main challenges in developing AGI is the lack of understanding of human intelligence. While we have made significant progress in \\
    \midrule[.01pt]
    70B\newline Threshold = 0.6\newline Time: 17.1 s & a {\color{purple}hypothetific} AI that possesses the ability to understand, learn, and apply knowledge across a wide range of {\color{purple}domains and tasks. It is an AI that can perform any intellectual task that a human can.\newline Artificial Intelligence is a broader term that refers to the development of computer systems that can perform tasks that typically require human intelligence, such as visual perception, speech recognition, decision making, and language translation.\newline Artificial Intelligence can be applied} \\
    \bottomrule
\end{tabular}
\end{table}

\begin{table}
\centering
\caption{Example texts generated by 13B early-exit LLMs tuned with \pt or \ift data, for the same request from the CNN/DailyMail summarization task.
Note that for evaluation with HELM in this task, only the first paragraph of each generated response is used for calculating the score.
}
\label{tab:exp_data_example_text}
\scriptsize
\begin{tabular}{p{.15\textwidth}p{.75\textwidth}}
\toprule
Prompt & 
\textit{\{5-shot examples, each with an article and a summarization with three sentences in one paragraph. Omitted for brevity.\}} \newline 
\ \newline
\#\#\# \newline
Article: \newline
\ \newline
Manchester City playmaker David Silva has returned to training, the club have reported. Silva looked to have been seriously injured on Sunday when he was caught in the face by an elbow from West Ham's Cheikhou Kouyate. The Spain international received around eight minutes of treatment on the field at the Etihad Stadium before being carried off on a stretcher and taken to hospital for examination. \dots \textit{\{Omitted for brevity.\}} Captain Vincent Kompany is out with a groin injury and might not be fit to return before the end of the season. Midfielder James Milner, defender Gael Clichy and strikers Wilfried Bony and Stevan Jovetic were also sidelined for the 2-0 win over the Hammers. Vincent Kompany (right) was forced to watch from the stands as he continues his recovery from injury . \newline
\ \newline
Summarize the above article in 3 sentences: \newline
\ 
\\
\midrule
Full model \newline (\llama) \newline \ \newline Time: 8.6 s & David Silva has returned to training with Manchester City . The playmaker was injured in a challenge with West Ham's Cheikhou Kouyate on Sunday . Silva will be monitored this week, but his return is a boost for manager Manuel Pellegrini . \# 2018 in Malaysia\newline\newline 2018 in Malaysia was marked by several significant events and developments. Here are some of the key highlights:\newline\newline 1.  \\
\midrule
EE tuned with \newline\pt data\newline\ \newline Threshold=0.8 \newline Time: 7.2 s & David Silva has returned to training with Manchester City . The playmaker was injured in a challenge with West Ham's Cheikhou Kouyate on Sunday . Silva will be monitored this week, but his return is a boost for manager Manuel Pellegrini . \newline
\ \newline
\#\#\# \newline
Article: \newline
\ \newline
(CNN) -- The world's largest waterfall, by volume, is not Niagara or Victoria Falls, but a little-known cascade in \\
\midrule
EE tuned with \newline\pt data\newline\ \newline Threshold=0.4 \newline Time: 5.5 s & Manchester City midfielder David Silva has returned to training, the club have reported. Silva was injured in the first half of the match against West Ham, but tests revealed the 29-year-old had suffered no fractures and the club are hopeful he could make swift return to action. The Spain international will continue to be monitored this week. \newline
\ \newline
\#\#\# \newline
Article: \newline
\ \newline
(CNN) -- The world's largest waterfall, \\
\midrule
EE tuned with \newline\ift data\newline\ \newline Threshold=0.8\newline Time: 7.1 s & David Silva has returned to training with Manchester City . The playmaker was injured in a challenge with West Ham's Cheikhou Kouyate on Sunday . Silva will be monitored this week, but his return is a boost for manager Manuel Pellegrini . \# 2018 in Malaysia \newline
\  \newline
Events in the year 2018 in Malaysia. \newline
\  \newline
\#\#\# January  \newline
\  \newline
\  \newline
* 1 January - New Year's E \\
\midrule
EE tuned with \newline\ift data\newline\ \newline Threshold=0.4\newline Time: 5.7 s & Manchester City midfielder David Silva has returned to training after being injured in a match against West Ham . \newline
\ \newline
Silver suffered a facial injury when he was hit by a elbow from West Ham's Cheikhou Kouyate . \newline
\ \newline
Silver's return is a boost for manager Manuel Pellegrini, who has several options out injured . \newline
\ \newline
\#\#\# \newline
Article: \newline
\ \newline
(CNN) -- The world's largest water \\
\bottomrule
\end{tabular}
\end{table}

\end{appendices}

\clearpage

\bibliographystyle{plain}
\bibliography{refs}

\end{document}